\documentclass[runningheads]{llncs}
\usepackage[T1]{fontenc}
\usepackage{graphicx}
\usepackage[caption=false]{subfig}
\usepackage{booktabs}
\usepackage{multirow}
\usepackage{makecell}
\usepackage{amsmath}
\usepackage{multirow}
\usepackage{tabularx}
\usepackage{xcolor}
\usepackage[normalem]{ulem}
\usepackage{hyperref}
\hypersetup{
    colorlinks,
    linkcolor={red!50!black},
    citecolor={blue!50!black},
    urlcolor={blue!80!black}
}
\usepackage{siunitx}
\useunder{\uline}{\ul}{}

\DeclareMathOperator{\mode}{mode}
\DeclareMathOperator{\tpr}{TPR}
\DeclareMathOperator{\fpr}{FPR}
\DeclareMathOperator{\tp}{TP}
\DeclareMathOperator{\tn}{TN}
\DeclareMathOperator{\fp}{FP}
\DeclareMathOperator{\fn}{FN}
\DeclareMathOperator{\auroc}{AUROC}
\DeclareMathOperator{\aupr}{AUPR}

\begin{document}
\title{
A Framework for Uncertainty Quantification Based on Nearest Neighbors Across Layers}

\titlerunning{Uncertainty Quantification Across Layers}

\author{
  Miguel N. Font\orcidID{0009-0008-6508-9137} \and
  José L. Jorro-Aragoneses\orcidID{0000-0002-2917-3912} \and
  Carlos M. Alaíz\orcidID{0000-0001-9410-1192}
}
\authorrunning{M. N. Font et al.}

\institute{Escuela Politécnica Superior, Universidad Autónoma de Madrid, Spain \email{miguel.font@estudiante.uam.es,\{jose.jorro,carlos.alaiz\}@uam.es}}


\maketitle

\begin{abstract}
Neural Networks have high accuracy in solving problems where it is difficult to detect patterns or create a logical model. However, these algorithms sometimes return wrong solutions, which become problematic in high-risk domains like medical diagnosis or autonomous driving. One strategy to detect and mitigate these errors is the measurement of the uncertainty over neural network decisions. In this paper, we present a novel post-hoc framework for measuring the uncertainty of a decision based on retrieved training cases that have a similar activation vector to the query for each layer. Based on these retrieved cases, we propose two new metrics: Decision Change and Layer Uncertainty, which capture changes in nearest-neighbor class distributions across layers. We evaluated our approach in a classification model for two datasets: CIFAR-10 and MNIST. The results show that these metrics enhance uncertainty estimation, especially in challenging classification tasks, outperforming softmax-based confidence. 
\end{abstract}

\section{Introduction}

There are numerous complex problems where detecting patterns or building effective models presents significant challenges, such as object detection in images~\cite{Egmont2002}. Neural Networks (NNs) in general, and Deep Neural Networks (DNNs) in particular, have become highly effective models for addressing this type of tasks. In particular, the increased amount of available data allows training these models accurately, while the current computational power allows the use of more complex and larger architectures. As a result, DNNs offer good results in domains like weather prediction, autonomous driving, or robotics~\cite{Abiodun2018}. However, these models are not foolproof, so detecting incorrect solutions becomes critical in high-risk domains like medical diagnosis.

A starting point to detect potential prediction errors in DNNs is to estimate the uncertainty of their decisions. In particular, uncertainty measurements attempt to quantify trust or ambiguity in predictions emitted by the model. In the literature, there are many approaches for estimating this uncertainty in the case of NNs and DNNs~\cite{abdar_review_2021}.

We present a novel approximation which estimates the uncertainty based on the dynamics of the network's decision-making.
Specifically, this framework retrieves training data that exhibit similar behavior to a query prediction by analyzing how the model's internal representations evolve across layers. Based on this information, we introduce two novel uncertainty metrics:
\begin{itemize}
    \item \textbf{Decision Change (DC):} Tracks the stability of class predictions across layers by detecting whether the most dominant class among the retrieved training cases fluctuates as the sample progresses through the network.
    \item \textbf{Layer Uncertainty (LU):} Measures the ambiguity within each layer by computing the entropy of the retrieved class distribution to assess how mixed the sample's neighborhood is.
\end{itemize}

To evaluate the validity of our proposal, we have conducted tests with two image datasets: CIFAR-10 and MNIST. In this evaluation, we measured the performance of the proposed DC and LU metrics for uncertainty quantification in detecting correct and incorrect predictions, comparing them to the use of the softmax output as a baseline~\cite{hendrycks_baseline_2018}. This evaluation also includes the study of the optimal number of neighbors needed to get good estimations. Additionally, we have included an evaluation of the performance when combining both proposed metrics to estimate uncertainty. The results show that our contributions are promising as uncertainty measures. 
Furthermore, the analysis of the retrieved cases allows users to verify whether errors stem from mislabeled data (e.g., incorrect labels in the neighborhood) or from the model itself (e.g., neighborhoods exhibiting wrong patterns, suggesting overfitting or poor learning).

The rest of the paper has the following structure. Section~\ref{sec:related_work} presents some of the most outstanding work on uncertainty estimation. Section~\ref{sec:approach} details our proposed framework and uncertainty metrics. Section~\ref{sec:results} presents the experimental setup and results obtained. Finally, Section~\ref{sec:conclusions} provides a summary of our findings and discusses potential future research directions.
\section{Related Work}
\label{sec:related_work}

Uncertainty quantification in NNs has attracted significant attention due to its critical role in high-stakes applications. Researchers have proposed diverse methodologies to measure model uncertainties. These approaches choose one of the following uncertainties to estimate~\cite{kiureghian_aleatory_2009}: aleatoric uncertainty, which arises from the data distribution (e.g., bias, data scarcity, noise, class overlap), and epistemic uncertainty, which reflects limitations in the model itself (e.g., design, parameters, underfitting). We have focused our study on approaches to epistemic uncertainty. There are four main types of uncertainty measure methods based on the number (single or multiple) and the nature (deterministic or stochastic) of the used DNNs~\cite{gawlikowski_survey_2023}. Following this taxonomy, the approach proposed later in this work corresponds to a single deterministic method.

Single deterministic methods estimate the uncertainty of a prediction using a single forward pass and may incorporate external components. Direct uncertainty prediction methods, such as those proposed by Kendall \emph{et al.}~\cite{kendall2017}, modify loss functions to estimate heteroscedastic aleatoric uncertainty (uncertainty that depends on the inputs, with some inputs potentially having more noisy outputs than others). However, these methods require retraining and model-specific adjustments, limiting their adoption in existing systems. Post-hoc calibration techniques, like softmax confidence baselines~\cite{hendrycks_baseline_2018} or temperature scaling~\cite{Liang2017}, avoid retraining but rely solely on final-layer outputs, overlooking the uncertainty dynamics in intermediate layers. The literature also includes more complex approaches, such as Gaussian methods~\cite{liu2020simple,mukhoti2023deep,tan2023single}, input transformation techniques~\cite{thiagarajan2022single}, and Radial Basis Function networks~\cite{vanAmersfoor2020}. However, these methods do not analyze how the prediction evolves within the neural network.

Similar to the framework proposed in this paper, there are some uncertainty measure alternatives based on feature-space distances. The Mahalanobis distance~\cite{lee2018} measures deviations from class-conditional distributions in latent space. The main goal of this approach is to detect out-of-distribution samples. However, this method assumes normality in feature distributions and focuses on static layer representations, neglecting layer-wise uncertainty and decision-making evolution. Another solution proposed by Kaber \emph{et. al.}~\cite{Kabir_2023} uses similar training cases to construct uncertainty margins, but it depends on external models to learn error patterns. Others, such as~\cite{ramalho2020}, estimate density in representation space but analyze only the final layer, omitting critical insights into how uncertainty propagates through the network.

Our work addresses these gaps by introducing a framework that analyzes layer-wise neighborhood evolution in activation spaces. Using information retrieval in all layers, we propose two complementary metrics: decision change, which quantifies prediction stability across layers by tracking class fluctuations in local neighborhoods, and layer uncertainty, which measures ambiguity within each layer via entropy of retrieved class distributions. Our approach is entirely post-hoc and model-agnostic---requiring no retraining, architectural changes, or external models---while capturing granular uncertainty signals missed by final-layer or single-metric strategies. In addition, our uncertainty measures are more transparent,  unlike previous proposals, because users can review the retrieval information used in the uncertainty quantification process.
\section{Proposed framework}
\label{sec:approach}

\begin{figure}
    \centering
    \includegraphics[width=\linewidth]{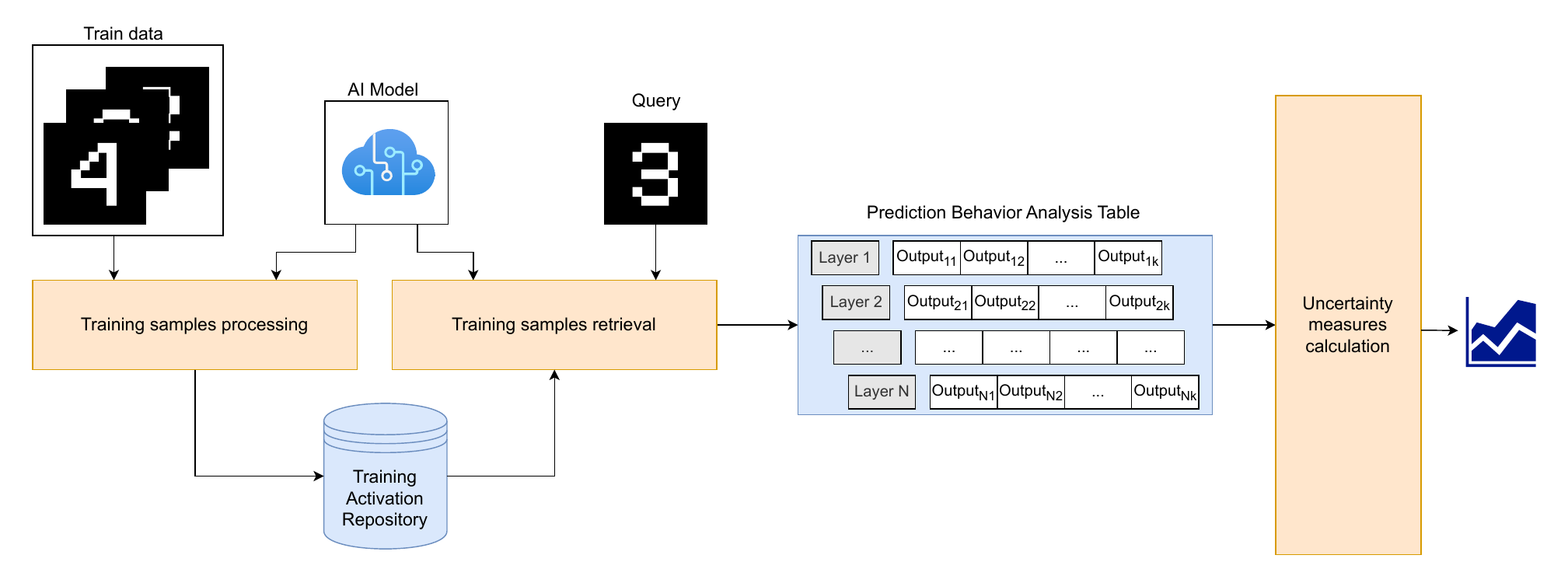}
    \caption{Overview of the proposed framework. The extraction of relevant cases is detailed in Subsections \ref{subsec:example_extract} and \ref{subsec:example_recovery}, while the definition and computation of uncertainty metrics is explained in Subsection \ref{subsec:metrics}.}
    \label{fig:Scheme_XAI_module}
\end{figure}

In this section, we explain our post-hoc approach to measuring uncertainty based on retrieving the nearest training samples in each layer. Figure~\ref{fig:Scheme_XAI_module} illustrates the entire framework process. It has three main steps. First, it processes the training samples to save their activation vectors in each of the layers, storing them in the training samples base. Second, when the NN models calculate a new prediction, the framework retrieves the most similar training samples for each layer and generates a table with the gathered information. Finally, the third step calculates the proposed uncertainty measures using this table. Each of these framework steps is detailed below.

\subsection{Training samples processing}
\label{subsec:example_extract}

This stage constructs a structured representation that enables the comparison of similar behaviors between a new prediction and the samples used to train the NN model. The framework processes all training samples through the model and stores the activation vectors generated at each layer. It then organizes this information into the \textit{Training Activation Repository} (TAR), which serves as a reference to retrieve the most similar training samples in the next stage.

\subsection{Training sample retrieval}
\label{subsec:example_recovery}

This step is triggered when a new prediction is emitted and its uncertainty has to be estimated. At that point, the framework requires the recovery of the most similar training samples stored in the TAR. In order to do so, it must store the activation vector of each layer of the NN for the prediction. Each of these activation vectors serves as a query to search within the TAR. Specifically, there is one query for each layer, and for each query, the framework retrieves the training samples that exhibit similar activation vectors at the corresponding layer. In our proposal, we use the Bray--Curtis distance to measure similarity, although our framework allows for the configuration of alternative distance metrics, e.g., Euclidean, Cosine, etc.

Once all the relevant information is retrieved, the framework organizes it into the \textit{Prediction Behavior Analysis Table} (PBAT), structuring the samples by layer. For each layer, the table stores the class label associated with each retrieved sample, as well as the identifier of the sample and the distance between its activation vector and the activation vector of the new prediction. This structure allows for the identification of the training samples, enabling their presentation to the end user in the form of a visualization if the data allows it. However, in this work, we have focused solely on using the class labels for further analysis.

\begin{figure}[!ht]
\centering
\subfloat{
    \raisebox{-.5\height}{\includegraphics[width=0.19\textwidth]{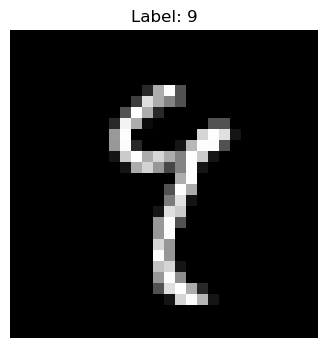}}
    \raisebox{-.5\height}{\includegraphics[width=0.48\textwidth]{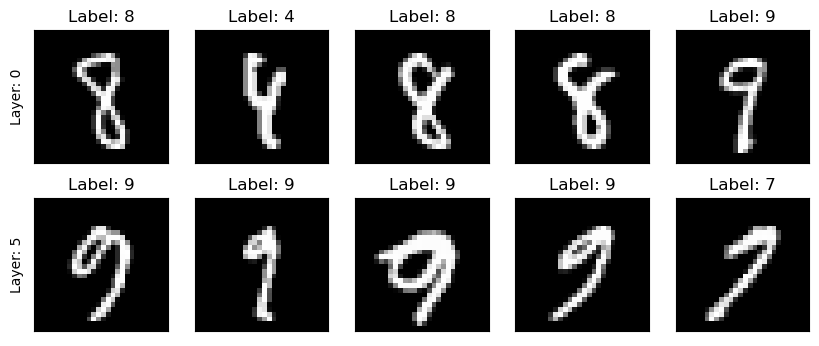}}
    \begin{tabular}{|c|p{0.35cm}p{0.35cm}p{0.35cm}p{0.35cm}p{0.35cm}|}
        \hline \textbf{Layer}  & \multicolumn{5}{c|}{\textbf{Closest cases}} \\ \hline
        1 & 8 & 4 & 8 & 8 & 9  \\
        \rotatebox{90}{$\cdots$} & \multicolumn{5}{c|}{\rotatebox{90}{$\cdots$}} \\
        5 & 9 & 9 & 9 & 9 & 7\\ \hline
    \end{tabular}
}
\quad
\subfloat{
    \raisebox{-.5\height}{\includegraphics[width=0.19\textwidth]{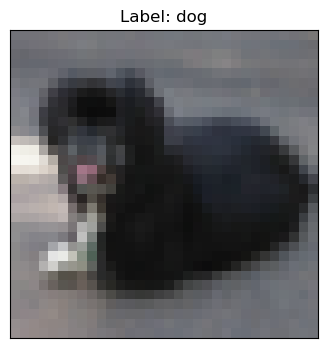}}
    \raisebox{-.5\height}{\includegraphics[width=0.48\textwidth]{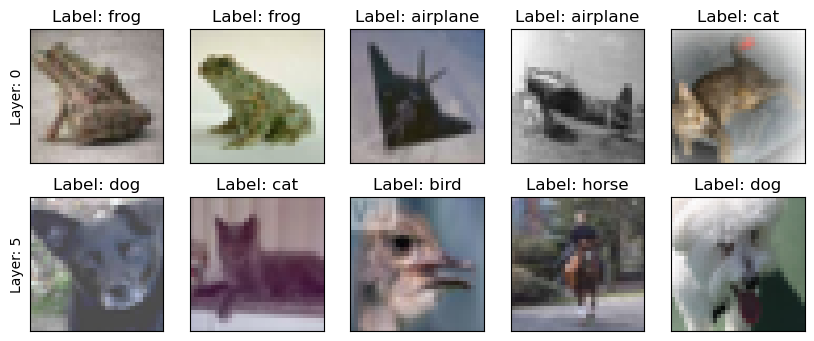}}
    \begin{tabular}{|c|p{0.35cm}p{0.35cm}p{0.35cm}p{0.35cm}p{0.35cm}|}
        \hline\textbf{Layer}  & \multicolumn{5}{c|}{\textbf{Closest cases}} \\ \hline
        1 & 6 & 6 & 0 & 0 & 3  \\
        \rotatebox{90}{$\cdots$} & \multicolumn{5}{c|}{\rotatebox{90}{$\cdots$}} \\
        5 & 5 & 3 & 2 & 7 & 5 \\\hline
    \end{tabular}
}
\caption{Five most similar samples and associated PBAT for the first and last layer (right) for both MNIST (up) and CIFAR-10 (down) datasets for a given input.}
\label{fig:sample_outputs}
\end{figure}

Figure~\ref{fig:sample_outputs} illustrates how this procedure works with a prediction example for both MNIST and CIFAR-10 datasets. On the left side, the new image to be classified is shown. In the central part, the five training samples retrieved from the \textit{TAR} for layers 0 and 5 of each dataset are displayed; as explained above, these are the instances with the most similar activation vectors to the prediction. On the right side, the \textit{PBAT} generated in both examples, based on the training samples recovered, is shown. This table will be used in the next phase to calculate the uncertainty measures proposed in this work.

\subsection{Uncertainty measures calculation}
\label{subsec:metrics}

Once the $k$ closest cases are identified for each layer, the framework computes two complementary uncertainty metrics that quantify different aspects of the behavior of the NN during inference. These metrics aim to capture two distinct aspects of the network's decision-making process: the stability of class predictions across layers (Decision Change) and the class ambiguity within each layer (Layer Uncertainty).

\subsubsection{Decision Change (DC):}

The \textit{Decision Change} metric quantifies the stability of the network's classification process as the sample propagates through the layers. It tracks how frequently the dominant class within the sample's local neighborhood changes from one layer to the next.

For each layer $\ell$, the dominant class is defined as the most frequent class label among the $k$ nearest cases at that layer:
\[
\hat{y}_\ell = \mode(\{y_i \mid i \in \mathcal{N}_\ell\}),
\]
where $\mathcal{N}_\ell$ denotes the set of the $k$ closest cases for layer $\ell$, and $y_i$ represents the true class label of the $i$-th case.

A \textit{decision change} occurs whenever the dominant class differs between two consecutive layers:
\[
\hat{y}_\ell \neq \hat{y}_{\ell-1}.
\]
Frequent decision changes suggest that the sample's internal representations are unstable, meaning that the network oscillates between classes as it refines its representation. These fluctuations can be a signal of potential misclassifications.

\subsubsection{Layer Uncertainty (LU):}

The \textit{Layer Uncertainty} metric quantifies the ambiguity present within the sample's neighborhood at each layer. It measures how evenly the $k$ nearest neighbors are distributed across different classes, providing a localized estimate of class confusion within the internal representation space.

Formally, for each layer $\ell$, this uncertainty is computed as the Shannon entropy of the class distribution among the $k$ nearest cases:
\[
H_\ell = -\sum_{c} p_\ell(c) \log p_\ell(c)
\]
where \(p_\ell(c)\) denotes the proportion of the $k$ nearest cases belonging to class $c$.

This metric is interpreted as follows:
\begin{itemize}
    \item \textbf{Low entropy} ($H_\ell \approx 0$): Most neighbors belong to a single class, indicating a high-confidence neighborhood with low ambiguity.
    \item \textbf{High entropy}: Neighbors are distributed across multiple classes, indicating that the sample lies in an ambiguous region of the feature space where the network struggles to assign a clear class, indicating low-confidence.
\end{itemize}

By computing this metric for each layer, the framework reveals how uncertainty evolves internally, potentially exposing ambiguous regions before the final classification is made. This offers complementary insight to the final softmax probability, which only reflects uncertainty at the very last step.
\section{Evaluation}\label{sec:results}

To evaluate the proposed framework and uncertainty metrics, we conducted an offline experiment with the goal of proving if correct and incorrect predictions are associated with our uncertainty measures. To this end, we divided the experiment into two stages. First, we searched for the optimal number of neighbors to retrieve for each scenario. Next, we evaluated the quality of the measurements and their combinations by comparing them with a baseline approach. We describe next the experimental setup and explain the obtained results.

All the code, experiments, and results explained in this section have been made available in a public online repository~\footnote{\url{https://anonymous.4open.science/r/Layer-Uncertainty-Framework-142C}}.

\subsection{Experimental setup}\label{subsec:exp_setup}

In this experiment, we have chosen two well-known datasets for image classification: \textbf{MNIST}~\cite{deng2012mnist} and \textbf{CIFAR-10}~\cite{Krizhevsky2009LearningML}. MNIST contains \num{70000} images of manuscript digits, and CIFAR-10 contains \num{60000} color images of \num{10} classes from objects and animals. Over both datasets, \num{10000} images were reserved as a test set.

The model used to classify both datasets was a Convolutional Neural Network (CNN) with two convolutional layers (\num{32} and \num{64} filters respectively, and each followed by a max-pooling layer), one fully connected layer (\num{64} neurons) and one final softmax classification layer (\num{10} neurons). The model was trained on each dataset for \num{5} epochs with an Adam optimizer at a learning rate of \num{0.001}.

We decided to use this experiment setup because it allowed us to test the framework on one setting with a high accuracy (\SI{98.39}{\percent} in MNIST) and another with a much lower accuracy (\SI{60.80}{\percent} in CIFAR-10).

To compare the usefulness of each uncertainty measure, we used simple linear models to classify the CNN predictions as correct or incorrect. In particular, we have built the following logistic regression models in this experiment: three that received as inputs only one uncertainty measure (softmax, DC or LU), one that received both DC and LU measures, and another where we used all three uncertainty measures as inputs.
We decided to use softmax confidence as a baseline, following the approach described by~\cite{hendrycks_baseline_2018}, where the predicted class' final probability can be used as a confidence estimate in classification tasks.
Notice that, since the proposed metrics provide layer-wise values, there will be as many features associated to DC and LU as layers has the NN---each metric provides a feature vector---.

The data used to train and evaluate each logistic regression model came from the test set of each dataset (MNIST and CIFAR-10). Each image in the test set was classified using the CNN, and we assigned a label indicating whether the classification was correct or not. We then split this new dataset, allocating \SI{80}{\percent} for training the logistic regression model and \SI{20}{\percent} for testing. To determine the optimal neighborhood sizes, we used \SI{20}{\percent} of the training data as validation data.

\subsection{Performance metrics}\label{subsec:perf_metrics}

In this experiment, our objective was to assess the usefulness of uncertainty measures in three cases: classifying both failures and correct classifications, classifying only correct classifications, and classifying only failures. To achieve this, we applied the following two metrics, as suggested by~\cite{hendrycks_baseline_2018}: \textit{Area Under the Receiver Operating Characteristic Curve} and \textit{Area Under the Precision-Recall Curve}.

\subsubsection{Area Under the Receiver Operating Characteristic Curve (AUROC)}

The Receiver Operating Characteristic (ROC) curve illustrates the trade-off between the \textit{True Positive Rate} ($\tpr$) and the \textit{False Positive Rate} ($\fpr$) across different decision thresholds of a classification model. These rates are defined as:
\[
\tpr = \frac{\tp}{\tp+\fn}, \quad \fpr = \frac{\fp}{\fp+\tn}.
\]

The area under the ROC ($\auroc$) summarizes this trade-off into a single scalar value, representing the model's ability to discriminate between correct and incorrect classifications and reflecting the probability that a randomly chosen correct classification receives a higher confidence score than a randomly chosen incorrect classification~\cite{FAWCETT}. In this context, an $\auroc$ value of \SI{50}{\percent} corresponds to random guessing, while a value of \SI{100}{\percent} indicates a perfect classifier.

\subsubsection{Area Under the Precision-Recall Curve (AUPR)}

While $\auroc$ is widely used, it can be misleading in cases where the positive and negative classes are highly imbalanced---a frequent situation in well-performing classifiers, where incorrect classifications are rare---. To address this, we also computed the \textit{Area Under the Precision-Recall Curve} ($\aupr$), which focuses directly on the classifier's performance in retrieving positive samples (correct classifications).

The \textit{Precision-Recall} (PR) curve plots:
\[
\operatorname{Precision} = \frac{\tp}{\tp+\fp}, \quad \operatorname{Recall} = \frac{\tp}{\tp+\fn} ,
\]
across varying confidence thresholds, capturing how well the model balances false positives and false negatives.

$\aupr$ directly measures the quality of the classifier's positive predictions, which is particularly relevant in our case, where identifying misclassifications (the minority class) is of primary interest. Notice that a simple baseline detector achieves an $\aupr$ approximately equal to the proportion of positive samples in the dataset~\cite{Saito2015}. 

To evaluate the model's performance in detecting correct and incorrect classifications, we applied two metrics based on the $\aupr$. The first one, $\aupr^{+}$, considers correct classifications as the positive class, with its baseline value corresponding to the network's overall accuracy. The second one, $\aupr^{-}$, treats incorrect classifications as the positive class, with a baseline value equal to the model's error rate (i.e., $1 - \operatorname{accuracy}$). In both cases, a perfect classifier would achieve an $\aupr$ of \SI{100}{\percent}.

\subsection{Results}
\label{subsec:results}

To test the performance of the proposed metrics, we start by choosing the optimal number of neighbors ($k$) for both datasets. The results of various configurations are shown in Table~\ref{tab:results_neighbors}.

\begin{table}[ht]
\centering
\caption{Quality of hit and miss detection over the validation set for different neighborhood sizes for DC and LU measures; each one is evaluated with three performance metrics ($\auroc$, $\aupr^{+}$ and $\aupr^{-}$) over both datasets (MNIST and CIFAR-10). Bold values mark the best result of the metrics in each dataset, while underlined values mark the second best result.}
\label{tab:results_neighbors}
\subfloat[Results for the DC measure.]{
\def\arraystretch{1.15}
\begin{tabular}{r@{\ }@{\ }c@{\ }@{\ }c@{\ }c@{\ }c@{\ }}
\toprule
 & $k$ & $\auroc$ & $\aupr^{+}$ & $\aupr^{-}$ \\ \midrule
\multirow{4}{*}{\rotatebox{90}{\textbf{MNIST}}}   & 3      & \textbf{84.65} & \textbf{99.57} & \textbf{17.0} \\
                         & 5                               & {\ul 80.39}    & {\ul 99.46}   & {\ul 15.6}    \\
                         & 10                              & 80.14          & 99.45         & 13.8          \\
                         & 20                              & 77.61          & 99.38         & 10.94          \\ \midrule
\multirow{4}{*}{\rotatebox{90}{\textbf{CIFAR-10}}} & 3     & 67.65          & 72.22         & 50.37          \\
                         & 5                               & 69.36          & 73.41         & 52.07    \\
                         & 10                              & {\ul70.49} & {\ul 74.09} & {\ul 53.52} \\
                         & 20                              & \textbf{71.38}          & \textbf{74.54} & \textbf{55.03} \\\bottomrule        
\end{tabular}
}\quad
\subfloat[Results for the LU measure.]{
\def\arraystretch{1.15}
\begin{tabular}{r@{\ }@{\ }c@{\ }@{\ }c@{\ }c@{\ }c@{\ }}
    \toprule
 & $k$ & $\auroc$ & $\aupr^{+}$ & $\aupr^{-}$ \\ \midrule
\multirow{4}{*}{\rotatebox{90}{\textbf{MNIST}}}   & 3      & 85.91          & 99.61             & 13.07                   \\
                         & 5                               & {\ul 90.81}          & {\ul 99.74}             & {\ul 17.64}             \\
                         & 10                              & 90.58    & {\ul 99.74}       & 16.43                   \\
                         & 20                              & \textbf{91.03} & \textbf{99.75}    & \textbf{19.05}          \\\midrule
\multirow{4}{*}{\rotatebox{90}{\textbf{CIFAR-10}}} & 3     & 67.91          & 72.29             & 50.84                   \\
                         & 5                               & {\ul 70.49}          & {\ul 74.24}             & {\ul 53.18}                   \\
                         & 10                              & 70.01    & 73.83       & 52.8             \\
                         & 20                              & \textbf{70.47} & \textbf{74.19}    & \textbf{53.22}          \\\bottomrule
\end{tabular}
}
\end{table}

We see that both measures are favored by a bigger neighborhood. This can be caused by the fact that more neighbors can provide more information about the uncertainty of the neighborhood. However, we observe an exception in DC for MNIST; this can be provoked because this dataset is simpler and more neighbors may keep the decision more stable across layers. Deeper study with more datasets should be carried out to better understand the influence of neighborhood sizes on the performance of the measures.

With the optimal number of neighbors selected, Table~\ref{tab:results} presents the results comparing the proposed uncertainty metrics (both individually and combined) against the softmax baseline.

\begin{table}[!ht]
\centering
\caption{Quality of hit and miss detection for each uncertainty measure over the test set against the dummy classifier (no algorithm) and the softmax (SM) uncertainty quantification baseline. Each measurement uses the optimal number of neighbors over the train set based on the results in Table~\ref{tab:results_neighbors} ($k=3$ for DC and $k=20$ for LU in MNIST, $k=20$ for DC and $k=20$ for LU in CIFAR-10).}
\label{tab:results}
\def\arraystretch{1.1}
\begin{tabular}{r@{\ }@{\ }l@{\ }@{\ }c@{\ }c@{\ }c@{\ }}
    \toprule
 & \textbf{Algorithm} & $\auroc$ & $\aupr^{+}$ & $\aupr^{-}$ \\ \midrule
\multirow{6}{*}{\rotatebox{90}{\textbf{MNIST}}}     & None & 50.00 & 98.39  & 1.61 \\
 & SM       & 90.03          & 99.60                & \textbf{33.96}      \\
 & DC       & 83.67          & 99.34                & 25.10               \\
 & LU       & 92.55          & {\ul 99.70}          & 24.75               \\
 & DC+LU    & {\ul 92.81}    & \textbf{99.71}       & 26.76               \\
 & SM+DC+LU & \textbf{92.93} & \textbf{99.71}       & {\ul 27.89}         \\ \midrule
\multirow{6}{*}{\rotatebox{90}{\textbf{CIFAR-10}}}  & None & 50.00 & 60.80 & 39.20 \\
 & SM       & 71.33         & 73.77                 & 55.50               \\
 & DC       & 71.13         & 73.41                 & 55.71               \\
 & LU       & 72.30         & 74.60                 & 56.37               \\
 & DC+LU    & {\ul 73.49}   & {\ul 75.59}           & {\ul 57.56}         \\
 & SM+DC+LU & \textbf{73.80} & \textbf{75.78}       & \textbf{57.99}       \\\bottomrule 
\end{tabular}
\end{table}

For MNIST, where the classification task is relatively easy, the softmax probability already achieves strong performance. However, combining the proposed metrics (DC and LU) improves the AUROC and AUPR for detecting errors, demonstrating that the internal uncertainty signals provide complementary information to the softmax confidence.

For CIFAR-10, a more challenging task with higher error rates, the combined metrics outperform the softmax baseline across all metrics, achieving higher AUROC and AUPR for both success and error detection. This suggests that the proposed framework is especially valuable in harder classification problems, where the softmax confidence alone may become less reliable.

Overall, the results confirm that the proposed metrics---particularly when combined---offer improved uncertainty estimation compared to the softmax baseline. This highlights the benefit of analyzing layer-by-layer neighborhood evolution, instead of relying solely on the final output probability. It is also noticeable that the proposed metrics do not need to be used alone and can be combined with the softmax uncertainty to further increase its capabilities, creating an effective tool for uncertainty quantification in both high and low accuracy regimes.
\section{Conclusion}
\label{sec:conclusions}

Neural Networks (NNs) are the state-of-the-art models in many real-life problems. Their popularity makes it critical to estimate the uncertainty of their predictions, especially in high-risk domains.
This paper proposes a layer-wise case-based reasoning framework for post-hoc Uncertainty Quantification (UQ) in NNs. By systematically tracking how nearest-neighbor class distributions evolve across the internal layers of a NN, a more detailed and interpretable view of uncertainty is enabled, revealing how ambiguity and instability emerge during the classification process.

In particular, two novel metrics can be defined under this framework: Decision Change (DC) and Layer Uncertainty (LU), which provide complementary insights into the classification process: DC reflects the stability of class assignments across layers, while LU quantifies class ambiguity at each layer. When used together, these metrics significantly enhance the detection of misclassified samples and provide better confidence estimates than using the default network accuracy or other uncertainty quantification methods such as the softmax probability.

The proposed framework is model-agnostic, requiring no retraining or architectural modifications, and can be applied to any trained feedforward network. Its flexible design, supporting different distance metrics, allows adaptation to different data domains and network types, which makes it both practical and generalizable.

A key limitation of the proposed framework is its computational overhead. During uncertainty quantification, the system must compute pairwise distances between each new input's activations and all cases stored in the training set. This can result in increased inference time, which could nevertheless be mitigated through several strategies, including case pre-filtering, approximate nearest neighbor search, or case condensation.

We acknowledge several promising directions to extend this work, such as using it for Out-of-Distribution (OoD) Detection, testing other distance metrics apart from Bray--Curtis, using it in other architectures like RNNs or transformers, combining it with other metrics like Mahalanobis distance and temperature scaling, or even generating visual explanations based on our layer-wise uncertainty proposal.

This work takes a step toward more interpretable and reliable post-hoc uncertainty quantification, bridging the gap between confidence estimation and explainability in neural networks.

\credits
\subsubsection*{Acknowledgements.}
This work was supported by project SI4/PJI/2024-00032 funded by the Comunidad de Madrid through the grant agreement, aimed at fostering and promoting research and technology transfer at the Universidad Autónoma de Madrid. The third author also acknowledge financial support from project PID2022-139856NB-I00 funded by MCIN/ AEI / 10.13039/501100011033 / FEDER UE, from Comunidad de Madrid e IDEA-CM (TEC-2024/COM-89).


\bibliographystyle{splncs04}
\bibliography{references}

\end{document}